\documentclass{article}

\PassOptionsToPackage{numbers}{natbib}

    \usepackage[preprint]{neurips_2023}

\usepackage[utf8]{inputenc} 
\usepackage[T1]{fontenc}    
\usepackage[colorlinks=true,linkcolor=blue,citecolor=teal]{hyperref}       
\usepackage{url}            
\usepackage{booktabs}       
\usepackage{amsfonts}       
\usepackage{nicefrac}       
\usepackage{microtype}      
\usepackage{xcolor}         
\usepackage{grffile}
\usepackage{subcaption}
\usepackage{arydshln}
\usepackage{tcolorbox}
\usepackage{listings}
\usepackage{enumitem}
\usepackage{adjustbox}

\newcommand{\chatgpt}{\texttt{gpt-3.5-turbo }}
\newcommand{\gpt}{\texttt{text-davinci-002 }}

\newtcbox{\inlinecodebox}{on line, colback=white, colframe=black, fontupper=\ttfamily, boxrule=0.5pt, boxsep=0pt, left=0.5pt, right=0.5pt,top=1pt,bottom=1pt}

\title{
Tool Documentation Enables Zero-Shot\\Tool-Usage with Large Language Models
}

\author{%
    Cheng-Yu Hsieh$^{1}$\footnotemark[2] , \
    Si-An Chen$^{2}$\footnotemark[2] , \
    Chun-Liang Li$^{3}$, \
    Yasuhisa Fujii$^{4}$, \\
    \bf
    Alexander Ratner$^{1}$, \
    Chen-Yu Lee$^{3}$, \
    Ranjay Krishna$^{1*}$, \
    Tomas Pfister$^{3*}$ \\
    $^1$University of Washington, \
    $^2$National Taiwan University,\\
    $^3$Google Cloud AI Research, \
    $^4$Google Research\\
    \texttt{cydhsieh@cs.washington.edu}
}

\begin{document}
\maketitle
\def\thefootnote{\dag}\footnotetext{Work done as student researchers at Google Cloud AI Research.}\def\thefootnote{\arabic{footnote}}
\def\thefootnote{*}\footnotetext{The authors contributed equally to this work.}\def\thefootnote{\arabic{footnote}}

\begin{abstract}
Today, large language models (LLMs) are taught to use new tools by providing a few demonstrations of the tool's usage.
Unfortunately, demonstrations are hard to acquire, and can result in undesirable biased usage if the wrong demonstration is chosen. 
Even in the rare scenario that demonstrations are readily available, there is no principled selection protocol to determine how many and which ones to provide. 
As tasks grow more complex, the selection search grows combinatorially and invariably becomes intractable.
Our work provides an alternative to \textbf{demonstrations}: tool \textbf{documentation}.
We advocate the use of tool documentation—descriptions for the individual tool usage—over demonstrations. 
We substantiate our claim through three main empirical findings on $6$ tasks across both vision and language modalities.
First, on existing benchmarks, zero-shot prompts with only tool documentation are sufficient for eliciting proper tool usage, achieving performance on par with few-shot prompts.
Second, on a newly collected realistic tool-use dataset with hundreds of available tool APIs, we show that tool documentation is significantly more valuable than demonstrations, with zero-shot documentation significantly outperforming few-shot without documentation.
Third, we highlight the benefits of tool documentations by tackling image generation and video tracking using just-released unseen state-of-the-art models as tools.
Finally, we highlight the possibility of using tool documentation to automatically enable new applications: by using nothing more than the documentation of GroundingDino, Stable Diffusion, XMem, and SAM, LLMs can \emph{re-invent} the functionalities of the just-released Grounded-SAM~\cite{groundedsam} and Track Anything~\cite{yang2023track} models.

\end{abstract}

\section{Introduction}
\label{sec:intro}

\begin{figure}[!t]
    \centering
    \includegraphics[width=0.95\linewidth]{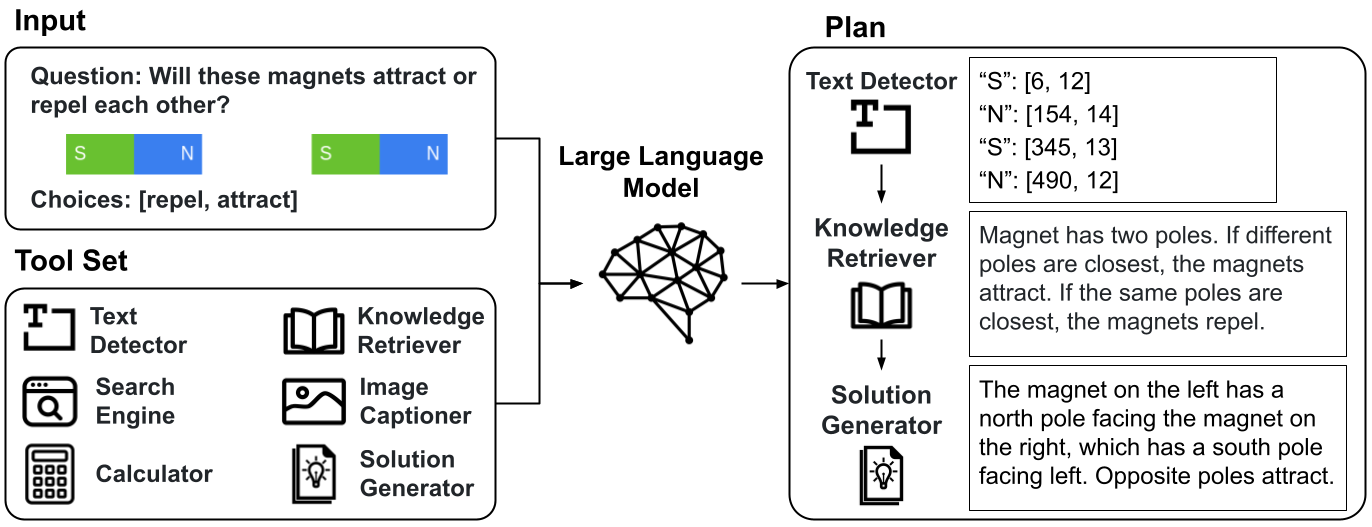}
    \vspace{-2mm}
    \caption{Example workflow of tool-using with LLMs to solve a multi-modal question answering task. Given the input question with an image, the LLM selects appropriate tools from the tool set and generates an execution plan to answer the question correctly. Here, the LLMs outlines a plan to first use Text Detector to understand the positioning of the magnets in the image, then leverage Knowledge Retriever to obtain relevant background knowledge about magnets, then finally generate the solution based on the previous steps.}
    \label{fig:tool_workflow}
    \vspace{-6mm}
\end{figure}

Today, large language models (LLMs) summon the imagery of a craftsman: when asked to solve a complex task, they decompose the task into simpler sub-tasks and assemble the best possible tools to tackle each sub-task~\cite{qin2023tool,yang2023foundation}.
For example, consider the complex task of question answering given the image in Figure~\ref{fig:tool_workflow}. To answer ``whether the two magnets will attract or repel each other'', the LLM needs the following: it needs to identify the positions of the magnets in the image, extract general knowledge explaining that ``opposite (same) poles attract (repel)''. Just like a competent craftsman who knows what their tools are capable of, an LLM with such knowledge of its tools will be able to invoke one tool (e.g.~its Text Detector) to identify the north and south poles and a second tool (e.g.~Knowledge Retriever) to extract pertinent background knowledge about magnetic forces.
But how does an LLM know which tool is capable of what?

\begin{figure}
    \centering
    \includegraphics[width=0.95\linewidth]{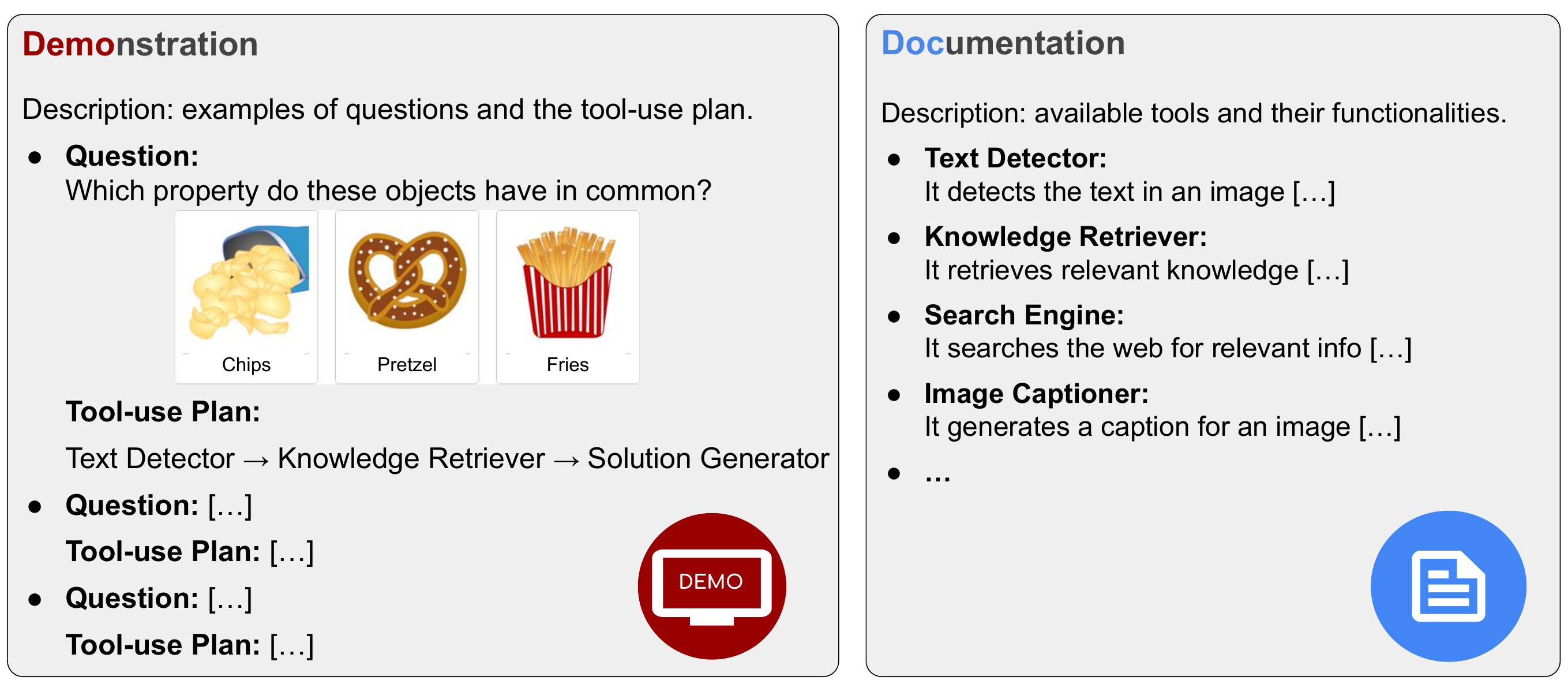}
    \vspace{-2mm}
    \caption{Two types of knowledge for prompting LLMs for tool-use: Demonstrations (demos) and Documentations (docs).
    Demos consist of <input, output> pairs on input instructions and their corresponding output tool-use plans. They require manual efforts for careful curation on every new task, and the model performance can be sensitive to which demos are used~\cite{zhao2021calibrate,Chen2023HowMD}. Many demos may also be necessary for good coverage when the number of tools scales up.  On the other hand, docs provide descriptions for the tool functionality, and are usually organically available for tools.
    }
    \label{fig:demo_vs_doc}
    \vspace{-10pt}
\end{figure}

Currently, LLM tool-usage provides LLMs with few-shot demonstrations (demos) of what its tools can do, hoping that these demos will help generalize the model's behavior to newer complex tasks. This process has been rather successful so far.
These few-shot demos contain one or several exemplars of <input, output> mappings~\cite{wei2023larger} on given instructions and their corresponding tool-use plans (illustrated in Figure~\ref{fig:demo_vs_doc}). LLMs are expected to find patterns within these demos and generalize them for new tasks.
On textual tasks, LLMs have presented with demos of calculators~\cite{cobbe2021training,parisi2022talm,schick2023toolformer}, Python interpreters~\cite{chen2022program,gao2022pal} and search engines~\cite{thoppilan2022lamda,nakano2021webgpt,press2022measuring,schick2023toolformer,lu2023chameleon} can perform logical and arithmetic operations to obtain more accurate and factual knowledge.
On visual tasks, LLMs with demos of pretrained vision models can do complex visual reasoning~\cite{liu2022mind,lu2023chameleon,shen2023hugginggpt,surismenon2023vipergpt,yang2023mm}, can generate and even edit images~\cite{gupta2022visual,bubeck2023sparks}.
On embodied robotic tasks, LLMs can similarly be used to reason and plan~\cite{yao2022react,huang2022language,saycan2022arxiv,driess2023palme}.

We argue that this reliance on demos in tool using is unnecessary in some cases, and might be even limiting.
In fact, recent work finds that LLMs tend to be sensitive to demos~\cite{zhao2021calibrate}, and carefully selecting demos is needed to avoid biasing or overfitting to a particular usage~\cite{Chen2023HowMD}.
This leads to the follow-up question: how do we choose which few-shot demos to use?
There are no known principled approaches to select demos without human intervention or to even efficiently enable humans to choose or create them.
To make the matter worse, when we scale up the number of tools that LLMs have access to, this few-shot selection process becomes combinatorially intractable. 
Just as a craftsman doesn't need to see a new tool being demonstrated and can instead discern their capabilities from reading a user manual for the tool, we seek to enable LLMs to learn how to use tools without seeing any demos.

Our work provides an alternative to \textbf{demonstrations}: tool \textbf{documentation} (doc).
Similar to the metaphor of a manual indicating an physical tool's capabilities, a software tool's docs outline what the tool can and cannot be used for and how to invoke it.
Docs provide relatively neutral instruction about the tools' functionalities and how individual tools should be used (illustrated in Figure~\ref{fig:demo_vs_doc}), and they are usually conveniently available through the creation of the tools organically.
Intuitively, just as the craftman leans to use a new tool by reading the manual, we provide LLMs with \texttt{README} files when encountering a new tool/repository.
With docs, an LLM may not necessarily need demos to use a new tool.

Distinct from existing work that rely mostly on few-shot demos for tool-learning, in this work, we study whether LLMs can instead solely rely on docs to use tools. 
We study the tool-learning performances of LLMs as we include or exclude docs, and vary the number of demos from few-shot down to zero-shot.
We conduct the experiments on $6$ tasks across vision and text modalities.
Our experiments show that:
\vspace{-3mm}
\begin{itemize}[leftmargin=20pt]
    \item Surprisingly, when provided with tool docs, LLMs' zero-shot tool-using performance is on par or even better than their few-shot counterparts, showing that including docs is an effective way to sidestep the few-shot demos needed.
    \item Building on the above finding, we relax the few-shot demo constraint, and show that we can efficiently scale up to a significantly larger tool set, on a newly collected API usage dataset, by simply providing the LLMs with docs.
    \item We show how to seamlessly add new tools along with their docs to a tool set for LLMs to solve unseen tasks on image editing and video tracking, all without any further demos in a plug-and-play manner.
    \item Finally, with unseen tools developed recently as building blocks, we showcase LLMs are capable of \emph{re-inventing} popular yet even more recent works Grounded-SAM~\cite{groundedsam} and Track Anything~\cite{yang2023track}, which suggests a potential from zero-shot tool usage to automatic knowledge discovery.
\end{itemize}

\section{Related work}
\paragraph{LLMs with retrieval augmentation and tools.}
In spite of the remarkable achievements demonstrated by LLMs, the performance can be further boosted with external tool usages to be more accurate, efficient or versatile for wider applications. 
The authors in~\cite{qin2023tool} detailed the cognitive origins, the paradigm shift of foundation models, and the complementary roles of tools and models to LLMs.
The example tool usage starts from knowledge retrieval~\cite{borgeaud2022improving,guu2020retrieval,lewis2020retrieval,yao2022webshop,yu2023generate} and expands to search engine~\cite{nakano2021webgpt,komeili2021internet,lazaridou2022internet,thoppilan2022lamda,shuster2022blenderbot,paranjape2023art,lu2023chameleon}, QA system~\cite{schick2023toolformer}, calculator~\cite{cobbe2021training,parisi2022talm,schick2023toolformer}, the Python interpreter~\cite{gao2022pal,chen2022program,wang2022code4struct,imani2023mathprompter,paranjape2023art,surismenon2023vipergpt}, simulation engines~\cite{liu2022mind}, machine learning models~\cite{shen2023hugginggpt,yang2023mm,wu2023visual,lu2023chameleon,surismenon2023vipergpt}, or even tools created by LLMs~\cite{cai2023large}.
Pioneer works of LLMs with tools often rely on human supervision~\cite{thoppilan2022lamda,komeili2021internet} or additional self-supervised learning techniques~\cite{schick2023toolformer}, which pose challenges for practical plug-and-play usage. 
Recent advancements eliminate additional training by using example demos in the prompt~\cite{gupta2022visual,yao2022react,yang2023mm,shen2023hugginggpt,lu2023chameleon,paranjape2023art}.
Our work further simplifies prompt design by only leveraging documentation for individual tools, while maintaining competitive performance.


\paragraph{Planning with LLMs.}
Language models are proven to have potential to conduct planning for solving complex tasks or decompose the complex tasks into sub-problems when prompted properly. 
\cite{huang2022language, huang2022inner} retrieve demos at test-time with large knowledge space coverage to generate admissible actions.
\cite{khot2022decomposed} relies on pre-designed demos for task decomposition. 
Similarly, recent works of tool using with LLMs leverage the example demonstrations of solving examples tasks with a planning of tools~\cite{chen2022program,gupta2022visual,yao2022react,yang2023mm,shen2023hugginggpt,lu2023chameleon,paranjape2023art}.  
However, crafting demos of interactions between tools may be challenging in practice when the number of tools surges.
Concurrent work~\cite{patil2023gorilla,qin2023toolllm,yang2023gpt4tools} tackles the challenge by using strong LLMs such as GPT-4~\cite{gpt-4} to create large instruction-following datasets that cover diverse instructions and corresponding tool-use plans, typically through mechanisms like self-instruct~\cite{wang2022self}. The resultant datasets can then be used to finetune and equip other LLMs (e.g., LLaMA~\cite{touvron2023llama} and OPT~\cite{zhang2022opt}) the ability to use a large collection of tools for unseen instructions. On the other hand, our work showcases the potential for LLMs to utilize any unseen new tools by reading their tool docs.

\paragraph{Demonstration and Documentation.}
Learning from demonstration is popular in reinforcement learning~\cite{pomerleau1988alvinn,bain1995framework,ng2000algorithms,ross2011reduction}. 
\cite{brown2020language} propose the in-context learning algorithm for efficient and effective  downstream task adaptations through showing example demonstrations. Inspired by the success, most of existing LLM tool-using works rely on few-shot demonstration~\cite{chen2022program,gupta2022visual,yao2022react,yang2023mm,shen2023hugginggpt,lu2023chameleon,paranjape2023art}. 
However, \cite{Chen2023HowMD} show that having more example demonstration might counter-intuitively degrade performance, and a careful selection might be needed.
\cite{liu2021makes} proposes a retrieval method for demo selection, which implicitly requires a larger set of examples to be selected. 
Using documentation to improve algorithms is relatively under-explored. 
\cite{branavan2012learning,zhong2019rtfm} propose document reading algorithms for specific games. \cite{zhou2023docprompting} introduced DocPrompting, which employs a trained retriever on the given training data to boost code generation by retrieving relevant documents.
In this work, we take a step towards exploring the zero-shot tool planning in LLMs solely with the aid of documentation, and investigate a wide range of diverse tasks from language to vision domains.
While~\cite{wang2023plan,Lu2023MultimodalPP} showcase pure zero-shot planning capability of LLMs, they do not study either the tool usage or the unseen scenarios to the language models. 
ViperGPT~\cite{surismenon2023vipergpt} is a concurrent work, which focuses on visual programming in Python and uses function implementations and specifications as documentation.
Lastly, while AutoGPT~\cite{autogpt} provides several demos that showcase the LLM's capability of tool using through documentation reading, our study focuses on a systematic exploration ranging from real-world use cases to academic benchmarks.

\section{Experimental setup}
\subsection{General workflow}
We follow the general framework of tool-using with LLMs in~\cite{qin2023tool}, which encompasses many of the recent works~\cite{yao2022react,khattab2022demonstrate,gupta2022visual,shen2023hugginggpt,yang2023mm,wu2023visual,lu2023chameleon}.
Specifically, given a natural language instruction, an \emph{LLM planner} generates a \emph{program} to be sequentially executed where each step of the program may rely on using tools selected from a \emph{tool set}. After the program is generated, it is then executed by an \emph{environment} which finally returns the execution results.
Here, the \emph{program} extends beyond conventional coding practice~\cite{yin2017syntactic,rabinovich2017abstract,iyer2018mapping} and is more closely associated with automata theory~\cite{sipser1996introduction}: a set of instructions of automations (e.g. tools in our case). 
Therefore, the tool set can be libraries with specific programming languages (e.g. Python), or general computation with properly defined input-output, such as trained models, API calls, and beyond.

\subsection{Tool-use prompting methods}

As discussed in Section~\ref{sec:intro}, two main types of information are considered in prompting LLMs for tool-using plans: demonstrations (demos) and documentations (docs). Demos showcase how tool \emph{interactions} can accomplish \emph{specific} tasks, while docs describe individual tool functionalities without task-specific ties as shown in Figure~\ref{fig:demo_vs_doc}.
In the experiment, we explore combinations of including/excluding docs and demos in prompts, as well as varying numbers of demos.


\subsection{Evaluation tasks}
We conduct our experiments on 6 tasks across multiple modalities with a variety of tool sets. We describe the setup and the tool sets for each task below. Except for specific cases where it is explicitly specified, the LLM planner is ChatGPT (\texttt{gpt-3.5-turbo}).

{\bf Multi-modal question answering on ScienceQA.}
ScienceQA~\cite{lu2022learn} consists of multi-modal multiple-choice science questions that requires language and visual understanding as well as domain-specific knowledge to answer correctly.
On ScienceQA, we follow the setup used in Chameleon~\cite{lu2023chameleon} and employ the same tool set with 7 tools, such as the search engine and the image text detector.

{\bf Tabular math reasoning on TabMWP.}
TabMWP~\cite{lu2023dynamic} is a math reasoning dataset with various forms of tables. It requires a model to understand structured or domain-specific tables, and utilize the information to answer corresponding math questions. On TabMWP, we also follow Chameleon~\cite{lu2023chameleon} with the same tool set with 9 tools, such as program generator and column lookup.

{\bf Multi-modal reasoning on NLVRv2.}
NLVRv2~\cite{suhr-etal-2019-corpus} requires the model to verify whether a statement is true on a pair of images, requiring compositional understanding of both texts and images. On NLVRv2, we follow the setup used in Visual Programming (VisProg)~\cite{gupta2022visual} with 20 vision modules (tools) for image understanding and manipulation.
Since VisProg only relies on few-shot demonstrations and does not utilize documentations for the modules. We generate the documentation for each module by including descriptions on the functionality of the module and the function signature. We provide the full documentations we use for each module in the appendix.

{\bf Unseen API usage on a newly collected dataset.}
Existing benchmarks used in literature come with a limited set of tools. To explore real-world use cases involving a large number of tools, we collect a new benchmark called the \emph{LLM Cloud CLI} that consists of 200 commands representing the functionalities of the Google Cloud Platform (GCP) command-line interface (CLI). Each command in our CLI is renamed from its corresponding GCP command, preserving the semantics and logic of the original tools, while being unseen to the language models.
For instance, the command \inlinecodebox{gcloud compute create NAME}, responsible for creating a virtual machine, is renamed to be \inlinecodebox{llmvm compute make NAME}. 
The renaming conventions also allow us to utilize authentic GCP examples as few-shot demos and leverage the corresponding GCP documentation. The benchmark comprises 50 questions, each focused on creating and configuring specific cloud services using  command-line tools. Each question requires at least two commands to complete the task. We show an example in Figure~\ref{fig:gcp_example}, and include more in appendix.

\begin{figure}[!t]
  \centering
  \includegraphics[width=0.99\textwidth]{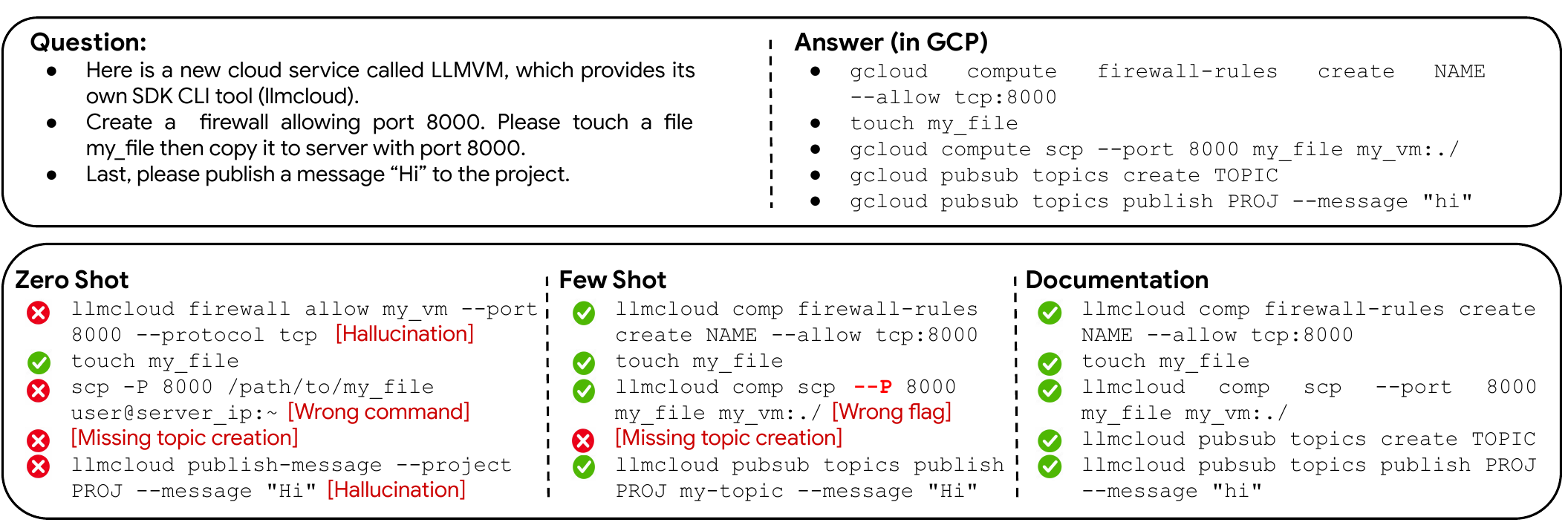}
  \vspace{-2mm}
  \caption{The new LLM Cloud Platform command-line toolkit, which is an unseen toolset to existing LLMs  based on real-world Google Cloud command-line tools through renaming.}
  \label{fig:gcp_example}
  \vspace{-4mm}
\end{figure}

Due to the length constraints of the LLM we use, we cannot fit documentation of 200 tools in a single prompt.
Therefore, we employ a simple TF-IDF search using the questions as queries to retrieve the most relevant documentations and truncate them to fit within the prompt length. More details can be found in the appendix.


{\bf Image editing with natural language.}
We consider image editing as a form of qualitative evaluation. This process calls for the model to plan and use different vision modules to handle complex natural language instructions.
For instance, to execute an instruction like "replace the red bus with a green bicycle", the model must localize the red bus, generate its segmentation mask, and then inpaint the masked area.
We use the tool sets from VisProg.
Unlike VisProg, which depends on few-shot demonstrations, our model only looks at the module documentation. 
We further include the recently released image understanding works, Segment Anything (SAM)~\cite{kirillov2023segment} and Grouding DINO~\cite{liu2023grounding} to expand the tool set to test the zero-shot capability on the new and unseen tools in a plug-and-play fashion.

{\bf Video tracking.}
Video tracking is also utilized in this study as a qualitative evaluation. This task aims to acquire the masks of a tracked object in each frame of a video, necessitating the deployment of processes such as object localization, segmentation, and tracking. In addition to SAM and Groudning DINO, we incorporate the documentation of an unseen object tracking module, Xmen~\cite{cheng2022xmem} into the VisProg framework with the aim to showcase the model's ability to adapt and employ new tools without the need for explicit demonstrations again on a different task.

\section{Empirical findings}
We showcase the importance of tool documentation in three-fold:
First, we show that tool documentations reduces the need of demonstrations (Section~\ref{sec:exp-chameleon}).
Second, based on the finding, we further show that relying on documentation rather than demonstrations provides a more scalable solution to equip LLMs with a large number of available tools (Section~\ref{sec:exp-gcp}).
Finally, we show that with tool documentations alone, LLMs are able to comprehend and utilize most recent vision models to accomplish impressive results on image editing and video tracking tasks, on which existing results are achieved either with human-crafted demos or predefined procedures (Section~\ref{sec:exp-plugnplay}).

\subsection{Documentations sidestep the need for demonstrations}
\label{sec:exp-chameleon}
In this section, we show how tool documentations reduce the need of demonstrations. We present the findings on three datasets: ScienceQA, TabMWP, and NLVRv2.
We evaluate the model performance, with and without tool documentations, across varying number of demonstrations (demo) on each dataset.

\begin{figure}[!t]
    \centering
    \includegraphics[width=0.98\linewidth]{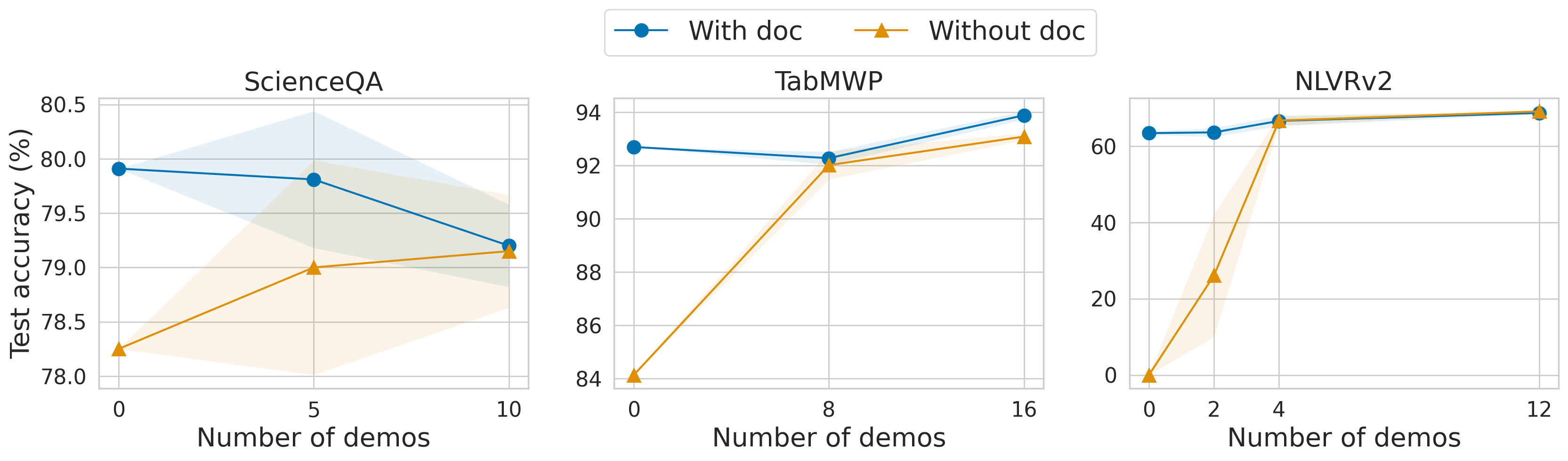}
    \vspace{-2mm}
    \caption{Tool-using performance with~\chatgpt on different benchmarks, which covers from langauge to vision modalities. We report results with and without documentation (doc) and demonstations (demo), and their combinations. Clearly, with documentation only (upper-left blue dot) shows competitive performance across all datasets.}
    \label{fig:doc_nodoc}
    \vspace{-3mm}
\end{figure}

In Figure~\ref{fig:doc_nodoc}, we see that when provided with tool docs, the model is able to maintain stable performance as we strip away the number of demos used. In fact, without using any demos (i.e., $0$-shot), the model is able to achieve on par performances to using $16$-shot on TabMWP, and using $12$-shot on NLVRv2. On ScienceQA, the model can even achieve better performance solely with docs compared to additionally using $10$-shot demos.
On the other hand, without tool docs, the model performance is very sensitive to the number of demos used. As we decrease the number of demos, we see significant performance drop on all three datasets.
This highlights the importance of tool docs and shows that it provides an effective way to reduce the reliance on demos. In Table~\ref{table:other_baselines}, when compared to existing baseline methods, we also see that with doc, even $0$-shot can perform very competitively.

By sidestepping the need for demos, we are able to alleviate the efforts needed to carefully curate these demos. For example, aligned with recent studies~\cite{zhao2021calibrate,Chen2023HowMD}, we observe in Figure~\ref{fig:doc_nodoc} that the model performance is sensitive to which demos are used, shown by the large performance variances under $5$-shot on ScienceQA and $2$-shot on NLVRv2.


\begin{table}[!t]
\centering
\caption{Comparisons to existing baseline methods on different benchmarks. We follow~\cite{lu2023chameleon,gupta2022visual} to select the beasline methods for each benchmark task. We see that $0$-shot with doc performs competitively, outperforming CoT and PoT on ScienceQA and TabMWP. On NLVRv2, ViLT-NLVR is finetuned on the dataset, while the LLM performs in a zero-shot fashion.}
\label{table:other_baselines}
\begin{tabular}{lccc}
\toprule
Benchmark & \multicolumn{3}{c}{Methods} \\
\midrule
 & CoT~\cite{wei2022chain}& without doc ($0$-shot) & with doc ($0$-shot) \\
\cmidrule{2-4}
ScienceQA & 78.54 & 78.25 & \textbf{79.91} \\
\midrule
 & PoT~\cite{chen2022program} & without doc ($0$-shot) & with doc ($0$-shot)\\
\cmidrule{2-4}
TabMWP & 89.28 & 84.13 & \textbf{92.69} \\
\midrule
 & ViLT-NLVR~\cite{kim2021vilt} & without doc ($0$-shot) & with doc ($0$-shot) \\
\cmidrule{2-4}
NLVRv2 & \textbf{76.30} & 0.00 & 63.40 \\
\bottomrule
\end{tabular}
\end{table}

\subsection{Documentations enable efficient scaling on tool-using}
\label{sec:exp-gcp}


The findings in Section~\ref{sec:exp-chameleon} show that one can in fact reduce the reliance on few-shot demos with tool docs. By relaxing this constraint, we study whether tool docs enables a more scalable way to equip LLMs with a large number of tools, wherein few-shot demos can specifically fall short on covering limited tool-use cases.
We present our findings in this section on the newly collected LLM Cloud CLI dataset with 200 available tools.

\paragraph{Qualitative walk-through result.}
Figure~\ref{fig:gcp_example} serves as a qualitative example illustrating the limitations of the LLMs with different information. As expected, zero-shot LLM successfully identifies and responds to the \inlinecodebox{touch} command, which is familiar and well-known. However, when faced with the unseen LLM-Cloud command lines, the zero-shot LLM fails to generate accurate responses involving these unfamiliar tools due to its lack of knowledge regarding their syntax and usage.


While few-shot demonstrations have the potential to enhance model performance, it is important to acknowledge that the coverage of these demonstrations is limited due to the vast number of command-line tools. Consequently, certain commands or flags may not be adequately covered. In Figure~\ref{fig:gcp_example}, although we observe  data copying is commonly appeared the few-shot examples, however, the model encounters difficulties in correctly configuring the less common flag \inlinecodebox{-{}-port}, instead hallucinating the use of \inlinecodebox{-P} based on familiarity with the \inlinecodebox{scp -P} command in Linux.

Conversely, in the same example illustrated in Figure~\ref{fig:gcp_example}, by solely utilizing the provided documentation, the language models not only successfully discern the steps required for utilizing tools (such as a hidden step of creating a \inlinecodebox{topic} before sending messages), but also possess the ability to accurately configure flags (e.g., \inlinecodebox{-{}-port}) by leveraging information extracted from the documentation.

\begin{figure}[t]
    \centering
    \includegraphics[width=.8\linewidth]{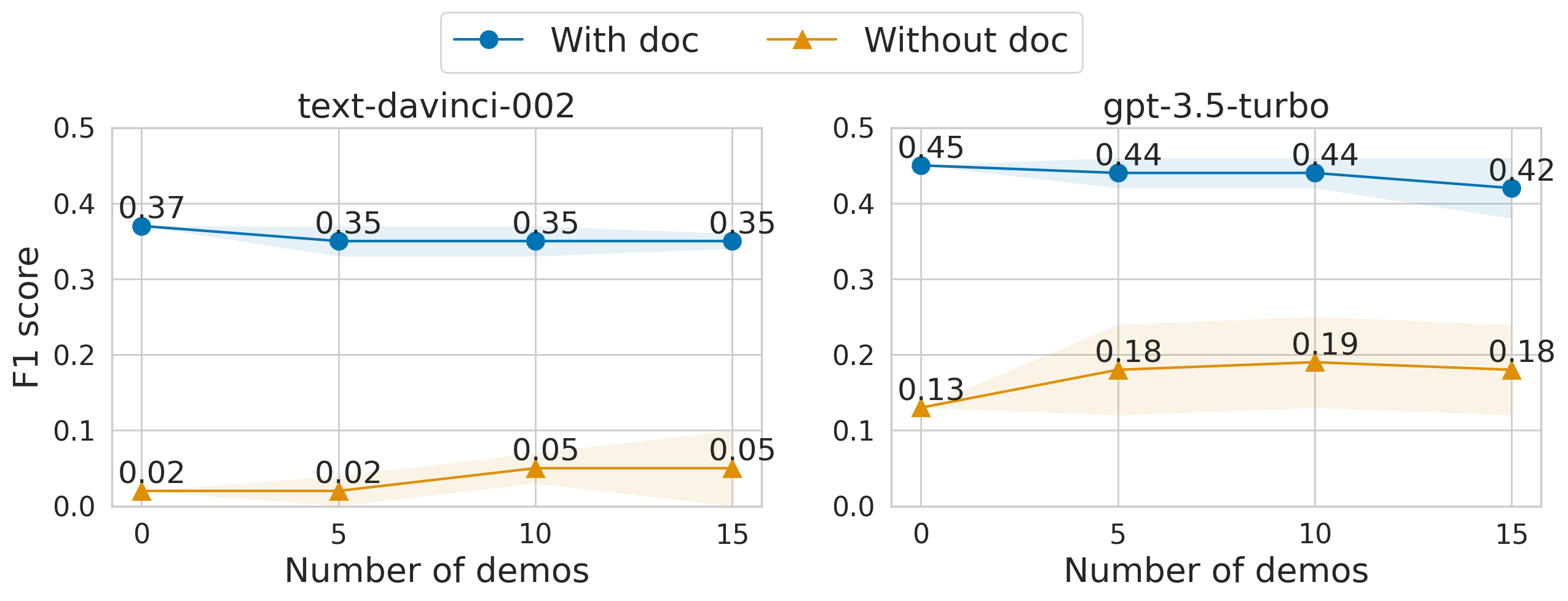}
    \vspace{-8pt}
    \caption{Command planning of LLM Cloud Platform CLI with and without documentation (doc) and demonstations (demo), and their combinations. Few-shot demonstration without documentation results in unsatisfactory performance due to low coverage of large number of tools, while reading documentation significantly boosts the performance.}
    \label{fig:gcp_doc_nodoc}
\end{figure}

\paragraph{Quantitative comparisons.}
We calculate the command-line level F1 score of each example and report the average F1 across 50 examples. 
Figure~\ref{fig:gcp_doc_nodoc} showcases the performance of various LLMs in the zero-shot setting, where they have no prior exposure to the LLM-Cloud command-line tools we create. As anticipated, all zero-shot LLMs demonstrate low F1 scores. Zero-shot \gpt achieves an F1 score of 0.02, while the \chatgpt model achieves a slightly higher score of 0.13. The improved performance of the \chatgpt model can be attributed to better handling of common Linux commands, such as \inlinecodebox{touch}.
As mentioned in quantitative comparison, few-shot demos improve upon zero-shot, but still fail on uncovered commands or flags in the demo. Therefore, the best few-shot demo in \gpt and \chatgpt are only with 0.05 and 0.19 F1 scores respectively.
On the other hand, LLM with documentation boosts the performance by a large margin to be 0.37 in \gpt and 0.45 in  \chatgpt.

We further compare the performance of the documentation reading with that of the documentation supplemented with few-shot demonstrations. In the case of \gpt, with documentation only, we achieves an F1 score of 0.37. Conversely, the documentation augmented with different shots yields an average F1 score of 0.35. 
Similarly, in the \chatgpt experiment, the performance with different shot demonstrations (0.44, 0.44, 0.42) are  consistently lower than the documentation-only performance (0.45).



These results highlight two observations. First, the performance of the model is highly sensitive to the selection of few-shot demonstrations.  The observation aligns the finding in ~\cite{Chen2023HowMD} that more few-shot demos might be redundant and even degrade performance due to spurious correlations.
It emphasizes the importance of careful selection and design, which may involve more human effort.
Second, the zero-shot documentation reading baseline exhibits remarkable robustness and delivers competitive performance across both examples. This highlights the potential value and reliability of relying solely on the documentation, which is usually easy to get in many packages and tools. 

\begin{figure}[t]
  \centering
  \includegraphics[width=\textwidth]{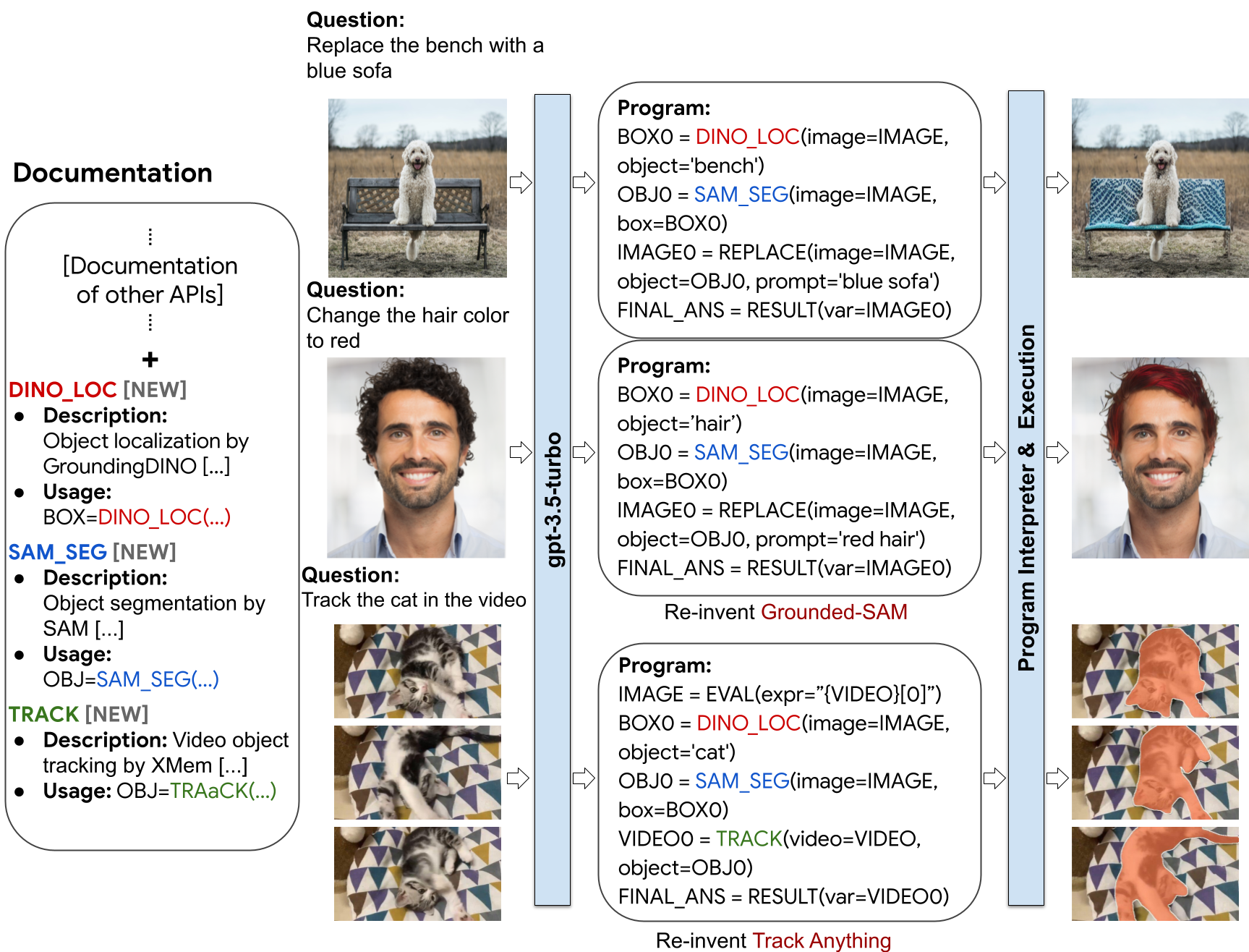}
  \vspace{-14pt}
  \caption{Plug-and-play new vision tools without demonstration. We add GroundingDINO~\cite{liu2023grounding}, Segment Anything (SAM)~\cite{kirillov2023segment}, XMem~\cite{cheng2022xmem} as new tools for VisProg. Solely with the documentations of the new tools, the LLM is able to automatically ``re-invent'' recent Grounded-SAM~\cite{groundedsam} and Track~Anything~\cite{yang2023track} without knowing these derivatives, taking a further step toward automatic knowledge discovery.}
  \label{fig:visprog}
\end{figure}

\subsection{Plug-and-play with new image and video tools}
\label{sec:exp-plugnplay}

In this section, we validate that one can equip LLMs with unseen tools to solve novel tasks solely with tool docs, and without any further demos. We present our results on image editing and video tracking tasks. We show that LLMs can effectively re-invent existing human-programmed image editing and video tracking pipelines, backed by state-of-the-art vision models to achieve impressive results.

Recent advancements in vision models, including GroundingDINO~\cite{liu2023grounding}, an advanced open-set object detector; Segment Anything (SAM)~\cite{kirillov2023segment}, a cutting-edge image segmentation tool; and XMem~\cite{cheng2022xmem}, a state-of-the-art video object segmentation tool, accompany the progress of language models. These breakthroughs, emerging in the past year, serve as additional tools that are yet unfamiliar to our LLM (\texttt{gpt-3.5-turbo}). By expanding VisProg to include these new tools, we embark on the intriguing exploration of whether LLMs can effortlessly comprehend the documentation associated with these new models, and combine these tools in a plug-and-play manner, enabling a wide range of applications.



In Figure~\ref{fig:visprog}, when performing an image editing request ``replace the bench with a blue sofa'', the LLM generates a VisProg program that harnesses the power of GroundingDINO and SAM from the expanded tool set to segment the bench, and apply the stable diffusion~\cite{rombach2022high} for synthesizing the sofa. This program \emph{re-invents the wheel} by replicating the behavior of recent popular project, Grounded-SAM~\cite{groundedsam} without prior knowledge of this repository. Similarly, when tasked with video tracking ``track the cat in the video'', the generated VisProg program by the LLM incorporates GroundingDINO together SAM for first frame segmentation as the initialization for XMem to do video tracking.
It again re-invents the results obtained in the contemporary work, Track Anything~\cite{yang2023track}. We note that TaskMatrix~\cite{wu2023visual} also has an updated approach with Grounded-SAM. However, they pre-program the entire Grounded-SAM editing pipeline as an image editing function, allowing the LLM to control it rather than enabling the LLM to generate the editing program using the building tools alone as we present here. 

By successfully re-inventing the functionalities of Grounded-SAM and Track Anything without prior knowledge, solely relying on the available building blocks, the LLM demonstrates not only its capacity to effortlessly comprehend and combine new tools with documentation only but also highlights its potential for automatic knowledge discovery.  
It discovers new insights through leveraging its existing knowledge only without further demonstration. 


\subsection{Performance v.s. documentation quality}
\begin{figure}[htbp]
  \centering
  \includegraphics[width=0.8\textwidth]{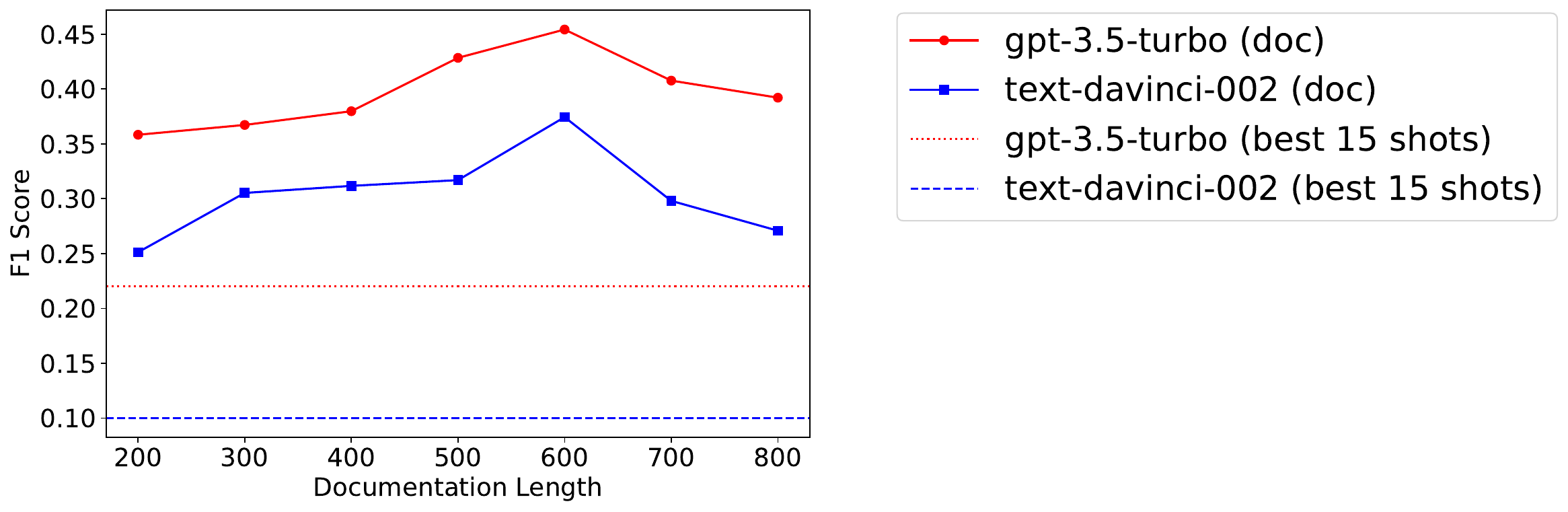}
  \vspace{-2mm}
  \caption{Performance of zero-shot documentation LLM when varying the input document length.}
  \label{fig:doc_len}
\end{figure}

We investigates the impact of documentation quality on performance. To assess LLM's capability to comprehend realistic documentation, we refrain from engineering or curating the content of the  documentation. Instead, we vary the document length by truncating the documents and keeping the first $n$ words, using it as a proxy for assessing thoroughness and quality.  In this ablation, we consider the LLM-Cloud benchmark, which has long documentation based on real-world GCP CLI manuals.
We illustrate the result in Figure~\ref{fig:doc_len}. 

In both \gpt and \chatgpt experiments, we consistently observe a trend where performance improves as the document length increases, up to a length of 600. 
This finding aligns with our hypothesis that the models possess the ability to comprehend and leverage documentation effectively. Remarkably, this improvement in performance is achieved without any additional training, fine-tuning nor document curation . It highlights the tremendous value of providing comprehensive documentation, as it empowers the models to leverage a wide range of command-line tools \emph{at scale}, solely through the process of reading and understanding the documentation.

We note that  a degradation in performance after the document length exceeds 600 words. We attribute this decline to the inherent challenges associated with comprehending lengthy documents in language models~\cite{sun2022chapterbreak}.
However, we foresee the ongoing advancements in handling long inputs in language models will gradually address this limitation \cite{bulatov2023scaling,bertsch2023unlimiformer,anthropic}. We leave exploring  solutions for overcoming this limitation for future research.
\section{Conclusion}
In this paper, we examined the effectiveness of tool docs in enabling zero-shot tool usage with LLMs. 
We first showed that LLMs can achieve on par or better performance than their few-shot counterparts when provided with tool docs. 
We then scaled up to a significantly larger tool set on a newly collected API through docs only. 
By simply plugging in new tools along with their docs, LLMs are able to tackle unseen tasks in image editing and video tracking without further demos and replicate the functionalities of recent popular projects, suggesting a potential for automatic knowledge discovery. 
Overall, we shed light on a new perspective of tool usage with LLMs by focusing on their internal planning and reasoning capabilities with docs, rather than explicitly guiding their behaviors with demos. 

\bibliographystyle{plain}
\bibliography{main}

\clearpage
\appendix
\section{Broader impacts and limitations}
This work studies the importance of tool documentations in equipping LLMs with the ability to compose usages of a variety of tools to accomplish complex tasks. However, as discussed in~\cite{qin2023tool}, it is imperative to contemplate what tools should be made available to LLMs as well as how one should interpret and rely on the results obtained from the models. We envision tool documentations as a channel to guide LLMs in more safely using the tools, aligning with the original intended use of the tools.

\section{Implementation details}
In this section, we provide further implementation details on each task.
We conduct all our experiments on Debian GNU/Linux 10 machines with $40$GB A100 GPUs.

\subsection{ScienceQA}
\begin{figure}[h]
    \centering
    \includegraphics[width=.8\linewidth]{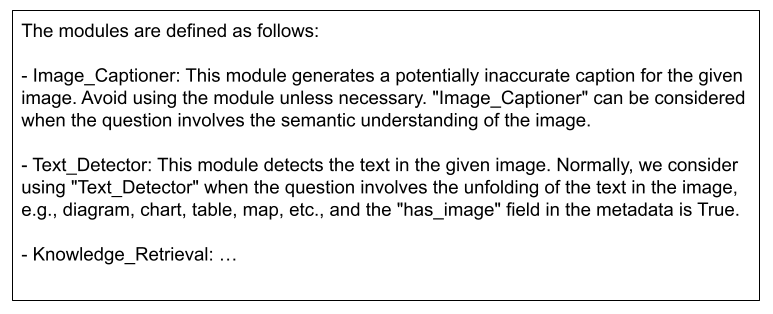}
    \caption{Documentations used in ScienceQA datasets. We used the original tool docs in Chameleon~\cite{lu2023chameleon} and added the description for ``Image Captioner'' that the generated captions may be inaccurate.}
    \label{fig:scienceqa_doc}
\end{figure}

On ScienceQA~\cite{lu2022learn}, we closely follow the original setup~\footnote{\url{https://github.com/lupantech/chameleon-llm}} used in Chameleon~\cite{lu2023chameleon}, including the tool docs and few-shot demos (when used). We however find that the ``Image Captioner'' module used in the original work often provides less accurate captions on given images. In the documentation, we thus add the description on this observation for the ``Image Captioner'' module as shown in Figure~\ref{fig:scienceqa_doc}.

\subsection{TabMWP}
On TabMWP~\cite{lu2023dynamic}, we strictly follow the original setup used in Chameleon~\cite{lu2023chameleon}. We refer the readers to~\cite{lu2023chameleon} and their open-sourced implementations for further details.

\subsection{NLVRv2}
\begin{figure}[!t]
    \centering
    \includegraphics[width=.9\linewidth]{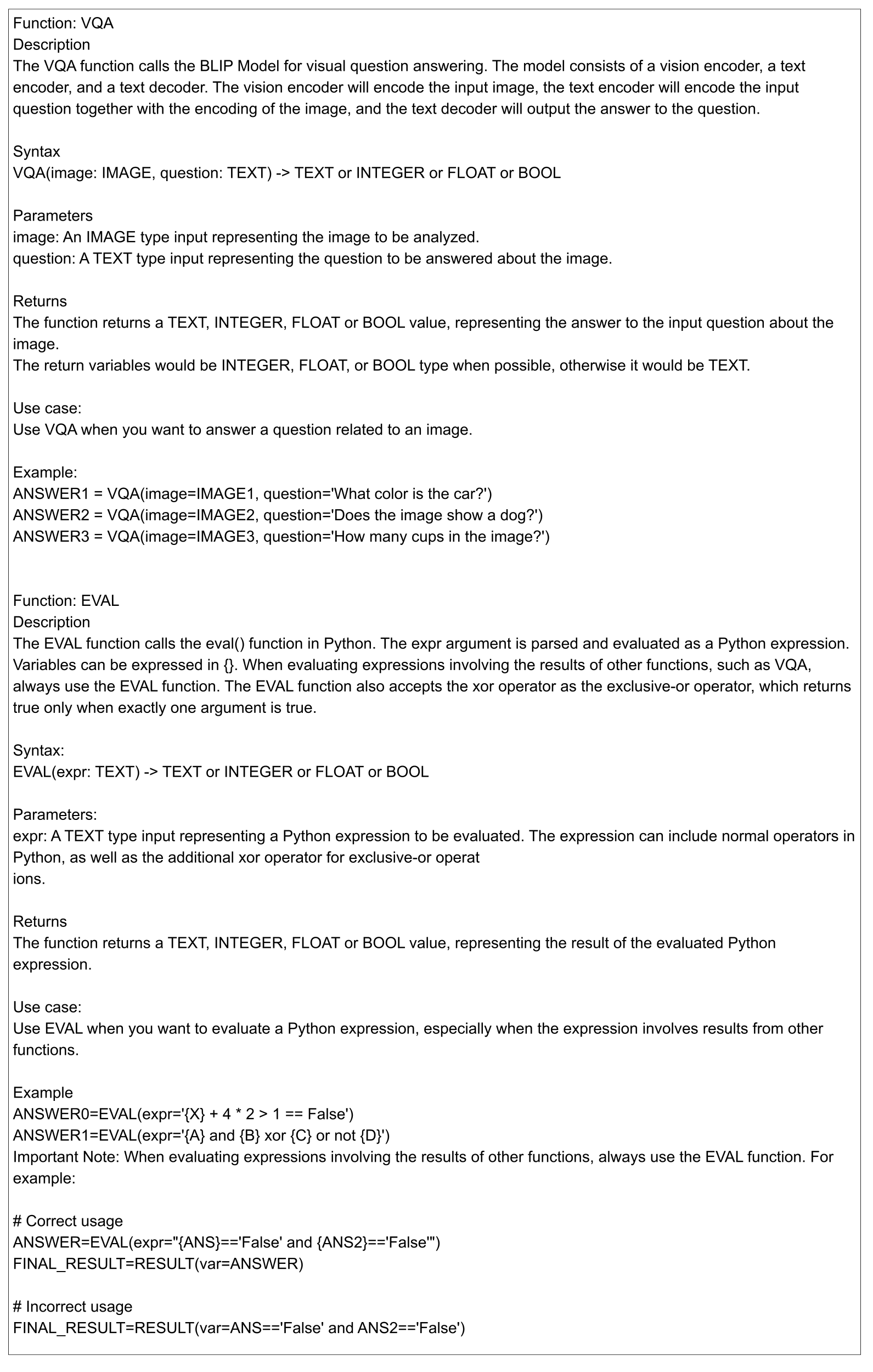}
    \caption{Example documentations used for tools in VisProg~\cite{gupta2022visual}.}
    \label{fig:nlvr_doc}
\end{figure}

On NLVRv2, we follow the setup used in~\cite{gupta2022visual}. However, as tool docs are not used in~\cite{gupta2022visual}, we create our own docs for the tools used. Figure~\ref{fig:nlvr_doc} shows the tool docs we use for several available tools used in VisProg~\cite{gupta2022visual}.

\clearpage

\subsection{LLM-Cloud CLI}
\paragraph{More examples.}
In Table~\ref{table:gcp_examples}, we show more examples of the created LLM-Cloud CLI dataset, based on GCP CLI.

\begin{figure}[!t]
  \centering
  \includegraphics[width=1.0\textwidth]{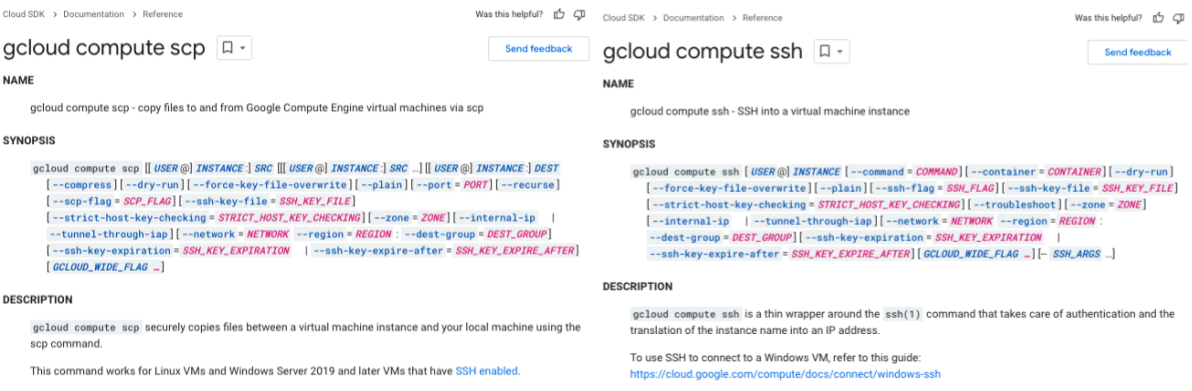}
  \caption{The documentation examples from GCP CLI. We crawl the website, remove the HTML tags and apply the renaming procedure as the documentation of the created LLM-Cloud CLI.}
  \label{fig:doc_example}
\end{figure}

\paragraph{Creating tool documentations.}
On the LLM-Cloud CLI dataset, we create tool documentations using the widely-used BeautifulSoup~\footnote{\url{https://pypi.org/project/beautifulsoup4/}} library to scrape the GCP CLI documentation. We removed HTML tags and implemented the renaming procedures for LLM-Cloud CLI documentation. We note that we purposely do not eliminate unrelated content such as terms and hyperlinks. An example documentation from GCP before our renaming procedure is shown in Figure~\ref{fig:doc_example}. This is to prevent excessive engineering of the documentation for better assessing the robustness of LLM documentation reading ability.

\paragraph{Documentation retrieval details.}
Given the extensive number of command-line tools in our experiments (200 in total), the complete documentation cannot fit within a single prompt. Consequently, for each query, we employ a simple TF-IDF search to retrieve the top 10 relevant documentations. We then truncate the length to a maximum of 600 words. We note that the actual token count depends on the tokenizer used by each LLM and is typically more than 600.

\begin{table}[!t]
\centering
\caption{More examples of the created LLM-Cloud CLI dataset.}
\label{table:gcp_examples}
\fontsize{9pt}{9pt}\selectfont
\begin{tabular}{|p{3.5cm}|p{4.75cm}|p{4.75cm}|}
  \hline
  \textbf{Question} & \textbf{Commands in GCP} & \textbf{Commands after renaming (Final Answer)} \\
  \hline
  Show me how to deploy ocr-xer container and invoke it with a schedule every 2 hours on a project ``test\_proj'' in sdk command lines. The ocr-xer container is located at ``us-docker.pkg.dev/gcr-cleaner/ocr-xer/ocr-xer''. 
  &
  \begin{itemize}[leftmargin=10pt]
    \item gcloud config set project test\_proj
    \item gcloud run deploy ocr-xer -{}-image=us-docker.pkg.dev/gcr-cleaner/ocr-xer/ocr-xer
    \item gcloud scheduler jobs create http NAME -{}-schedule -{}-schedule="0 */2 * * *"
  \end{itemize}
  &
  \begin{itemize}[leftmargin=10pt]
    \item llmcloud config set project test\_proj
    \item llmcloud run deploy ocr-xer -{}-image=us-docker.pkg.dev/gcr-cleaner/ocr-xer/ocr-xer
    \item llmcloud scheduler jobs make http NAME -{}-schedule -{}-schedule="0 */2 * * *"
  \end{itemize}
  \\
  \hline
  How to deploy a machine learning model model.pt saved in my local to cloud via sdk command line? 
  &
  \begin{itemize}[leftmargin=10pt]
    \item gsutil cp model.pt LOC/model.pt
    \item gcloud ai-platform versions create VERSION -{}-model MODEL -{}-origin gs://LOC/model.pt 
  \end{itemize}
  &
  \begin{itemize}[leftmargin=10pt]
    \item llmutil cp model.pt LOC/model.pt
    \item llmcloud ai-platform versions create VERSION -{}-model MODEL -{}-origin gs://LOC/model.pt 
  \end{itemize}
  \\
  \hline
  How to get transcript of a video test.mp4 at local via the cloud SDK?
  &
  \begin{itemize}[leftmargin=10pt]
    \item ffmpeg -i test.mp4 -ac 2 -f wav output.wav 
    \item gsutil cp test.wav LOC/test.wav
    \item gcloud ml speech recognize-long-running -{}-uri LOC/test.wav
  \end{itemize}
  &
  \begin{itemize}[leftmargin=10pt]
    \item ffmpeg -i test.mp4 -ac 2 -f wav output.wav 
    \item llmutil cp test.wav LOC/test.wav
    \item llmcloud ml speech recognize-long-running -{}-uri LOC/test.wav
  \end{itemize}
  \\
  \hline
  How to create a composer enviroment with a private ip network?
  &
  \begin{itemize}[leftmargin=10pt]
    \item gcloud composer environments create my\_env 
    \item gcloud compute networks subnets update default \
    -{}-enable-private-ip-google-access
  \end{itemize}
  &
  \begin{itemize}[leftmargin=10pt]
    \item llmcloud composer environments make my\_env 
    \item llmcloud compute networks subnets update default \
    -{}-enable-private-ip-google-access
  \end{itemize}
  \\
  \hline
  How to create a service account test@service.com with the name ``AutoML'' ``BigQuery Data Editor'' and  ``"AutoML Recommendations Service Account'' permissions?
  &
  \begin{itemize}[leftmargin=10pt]
    \item gcloud iam service-accounts test@service.com -{}-display-name AutoML
    \item gcloud projects add-iam-policy-binding PROJ\_ID -{}-member="test@service.com" -{}-role "roles/bigquery.dataEditor"
    \item gcloud projects add-iam-policy-binding PROJ\_ID -{}-member "test@service.com" -{}-role "roles/automlrecommendations.serviceAgent"
  \end{itemize}
  &
  \begin{itemize}[leftmargin=10pt]
    \item llmcloud iam service-accounts test@service.com -{}-display-name AutoML
    \item llmcloud projects add-iam-policy-binding PROJ\_ID -{}-member="test@service.com" -{}-role "roles/bigquery.dataEditor"
    \item llmcloud projects add-iam-policy-binding PROJ\_ID -{}-member "test@service.com" -{}-role "roles/automlrecommendations.serviceAgent"
  \end{itemize}
  \\
  \hline
\end{tabular}
\end{table}

\clearpage

\subsection{Image editing and video tracking}
\begin{figure}[h]
  \centering
  \includegraphics[width=.86\linewidth]{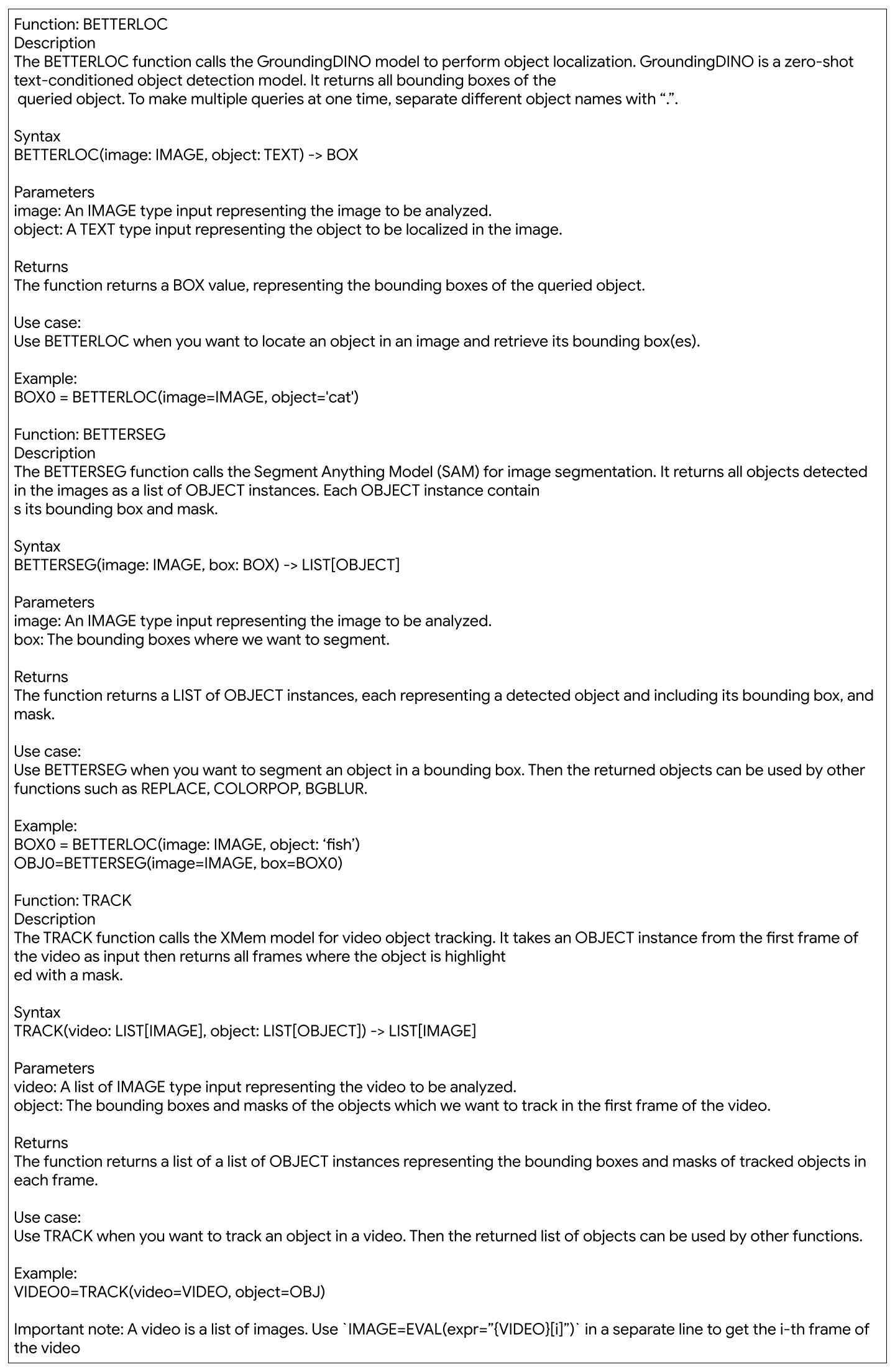}
  \caption{Documentation of new tools introduced in VisProg. BETTERLOC, BETTERSEG, TRACK calls GroundingDINO, Segment Anything, XMem, respectively.}
  \label{fig:visprog_doc}
\end{figure}

As discussed in Section~\ref{sec:exp-plugnplay}, by providing tool documentations, we can easily add on new tools to enable LLMs in solving novel tasks such as image editing and video tracking. Here, we leverage the recent advancements in vision models and expand the tool set used in VisProg~\cite{gupta2022visual} with three new tools: GroundingDINO~\cite{liu2023grounding}, Segment Anything (SAM)~\cite{kirillov2023segment}, and XMem~\cite{cheng2022xmem}. We provide their corresponding documentations in Figure~\ref{fig:visprog_doc}.

\clearpage

\section{Experimental results}
In this section, we show the experimental results on each task with comparisons to more baselines.

\paragraph{ScienceQA.} In Table~\ref{table:scienceqa}, we compare zero-shot prompting with tool documentations to other baseline methods. We include the following baseline methods that are finetuned on the ScienceQA training set for performance reference: ViLT~\cite{kim2021vilt}, VisualBERT~\cite{li2019visualbert}, UnifiedQA CoT~\cite{lu2022learn}, MM-CoT~\cite{zhang2023multimodal}, and LLaMA-Adapter~\cite{zhang2023llama}. We report the results obtained from~\cite{lu2023chameleon} for the finetuned methods.
For fair comparison, we shall focus on zero/few-shot settings. Thus, we include Chain-of-Thought (CoT)~\cite{wei2022chain} and Chameleon~\cite{lu2023chameleon} as the few-shot baselines to compare to.
We see that with tool docs, we can not only achieve better performance than the few-shot methods without any demos, but we can also match (outperform) several models specifically finetuned on the dataset.

\begin{table}[h]
\centering
\caption{Comparing zero-shot prompting with tool docs to existing baseline methods on ScienceQA. We see that zero-shot prompting with tool docs performs competitively, outperforming the two few-shot baselines and several finetuned models.}
\label{table:scienceqa}
\begin{adjustbox}{width=1\linewidth}
\begin{tabular}{lcccccccc}
\toprule
 & \multicolumn{5}{c}{Finetuned methods} & \multicolumn{2}{c}{Few-shot methods} & \multicolumn{1}{c}{Zero-shot methods}\\
\cmidrule(lr){2-6} \cmidrule(lr){7-8}  \cmidrule(lr){9-9}
Benchmark & ViLT & VisualBERT & UnifiedQA CoT & MM-CoT & LLaMA-Adapter &  CoT & Chameleon &  $0$-shot with docs \\
\midrule
ScienceQA & 61.14 & 61.87 & 74.11 & 84.91 & 85.19 & 78.54 & 79.20 & 79.91 \\
\bottomrule
\end{tabular}
\end{adjustbox}
\end{table}

\paragraph{TabMWP.}
Similarly, in Table~\ref{table:tabmwp}, we compare zero-shot prompting with tool docs to various finetuned models and few-shot baselines, inlcuding: UnifiedQA~\cite{khashabi-etal-2020-unifiedqa}, TAPEX~\cite{liu2021tapex}, Chain-of-Thought (CoT)~\cite{wei2022chain}, Program-of-Thought (PoT)~\cite{chen2022program}, and Chameleon~\cite{lu2023chameleon}. We report the results obtained from~\cite{lu2023chameleon} for UnifiedQA, TAPEX, and CoT.
We see that with tool docs, zero-shot prompting significantly outperforms finetuned models, and baseline few-shot methods, CoT and PoT. When compared to Chameleon that utilizes $16$ few-shot tool-usage demos, tool docs enable the model to perform comparably without relying on any demos.

\begin{table}[h]
\centering
\caption{Comparing zero-shot prompting with tool docs to existing baseline methods on TabMWP. We see that with tool docs, even zero-shot prompting without any tool-usage demos achieves better performance than finetuned models and few-shot CoT and PoT baseline. It also performs comparably to Chameleon that employs $16$-shot tool-usage demos.}
\label{table:tabmwp}
\begin{tabular}{lcccccc}
\toprule
 & \multicolumn{2}{c}{Finetuned methods} & \multicolumn{3}{c}{Few-shot methods} & \multicolumn{1}{c}{Zero-shot methods}\\
\cmidrule(lr){2-3} \cmidrule(lr){4-6}  \cmidrule(lr){7-7}
Benchmark & UnifiedQA & TAPEX & CoT & PoT & Chameleon &  $0$-shot with docs \\
\midrule
TabMWP & 57.35 & 58.52 & 82.03 & 89.28 & 93.88 & 92.69 \\
\bottomrule
\end{tabular}
\end{table}

\paragraph{NLVRv2.} In Table~\ref{table:nlvr}, we compare zero-shot prompting with tool docs to a finetuned model on NLVRv2 and various few-shot baselines. Specifically, we consider ViLT~\cite{kim2021vilt} as the finetuned baseline and VisProg~\cite{gupta2022visual} with varying numbers of tool-usage demos as the few-shot baselines. We report the result obtained from~\cite{gupta2022visual} for ViLT. Since VisProg does not utilize tool docs, we see that its performance is very sensitive to the number of demos used. In addition, we also observe large performance variances when we randomly select different demos used for prompting, e.g., the standard deviation for $2$-shot prompting reaches $16.1$ percentage point. This indicates that the few-shot demos may require careful curation for the model to achieve good performance.
On the other hand, with tool docs, zero-shot prompting can already achieve decent performance compared to only using few-shot demos.

\begin{table}[h]
\centering
\caption{Comparing zero-shot prompting with tool docs to existing baseline methods on NLVRv2.}
\label{table:nlvr}
\begin{adjustbox}{width=1\linewidth}
\begin{tabular}{lccccccc}
\toprule
 & \multicolumn{1}{c}{Finetuned methods} & \multicolumn{4}{c}{Few-shot methods} & \multicolumn{1}{c}{Zero-shot methods}\\
\cmidrule(lr){2-2} \cmidrule(lr){3-6}  \cmidrule(lr){7-7}
Benchmark & ViLT & VisProg ($0$-shot) & VisProg ($2$-shot) & VisProg ($4$-shot) & VisProg ($12$-shot) & $0$-shot with docs \\
\midrule
NLVRv2 &  76.30 & $0$ & $43.1 \pm 16.1$ & $66.5 \pm 1.4$ & $69.1 \pm 0.1$ & $63.4$\\
\bottomrule
\end{tabular}
\end{adjustbox}
\end{table}

\clearpage

\paragraph{LLM Cloud-CLI.} In Table~\ref{table:gcp}, we present the results on LLM-Cloud CLI with different underlying LLM planners. On both \texttt{text-davinci-002} and \texttt{gpt-3.5-turbo}, when there is a large number of tools, we see documentation is much more important than few-shot demonstrations, where zero-shot with docs achieves significantly better performances than few-shot without docs.
Additionally, when provided with docs, the LLMs are able to figure out how to use the tools without the need of demonstrations.

\begin{table}[!h]
  \centering
  \caption{Results on the  LLM-Cloud CLI.}
  \label{table:gcp}
  \begin{tabular}{cccc}
    \hline
    LLM & Number of Demos & Documentations & F1 \\
    \hline
    \texttt{text-davinci-002} & 0 & No & 0.02 \\
    & 5 & No & $0.02 \pm 0.02 (0.05)$ \\
    & 10 & No & $0.05 \pm 0.02 (0.11)$ \\
    & 15 & No & $0.05 \pm 0.05 (0.1)$ \\
    \hdashline[1pt/1pt]
    & 5 & Yes & $0.35 \pm 0.02 ({\bf 0.37})$ \\
    & 10 & Yes & $0.35 \pm 0.02 ({\bf 0.37})$ \\
    & 15 & Yes & $0.35 \pm 0.01 ({\bf 0.37})$ \\
    \hdashline[1pt/1pt]
    & 0 & Yes & {\bf 0.37} \\
    \hline
    \hline
    \texttt{gpt-3.5-turbo} & 0 & No & 0.13 \\
    & 5 & No & $0.18 \pm 0.06 (0.21)$ \\
    & 10 & No & $0.19 \pm 0.06 (0.23)$ \\
    & 15 & No & $0.18 \pm 0.06 (0.22)$ \\
    \hdashline[1pt/1pt]
    & 5 & Yes & $0.44 \pm 0.02 ({\bf 0.47})$ \\
    & 10 & Yes & $0.44 \pm 0.02 ({\bf 0.48})$ \\
    & 15 & Yes & $0.42 \pm 0.04 ({\bf 0.49})$ \\
    \hdashline[1pt/1pt]
    & 0 & Yes & {\bf 0.45} \\
    \hline
  \end{tabular}
\end{table}

\begin{figure}[!t]
  \centering
  \includegraphics[width=\linewidth]{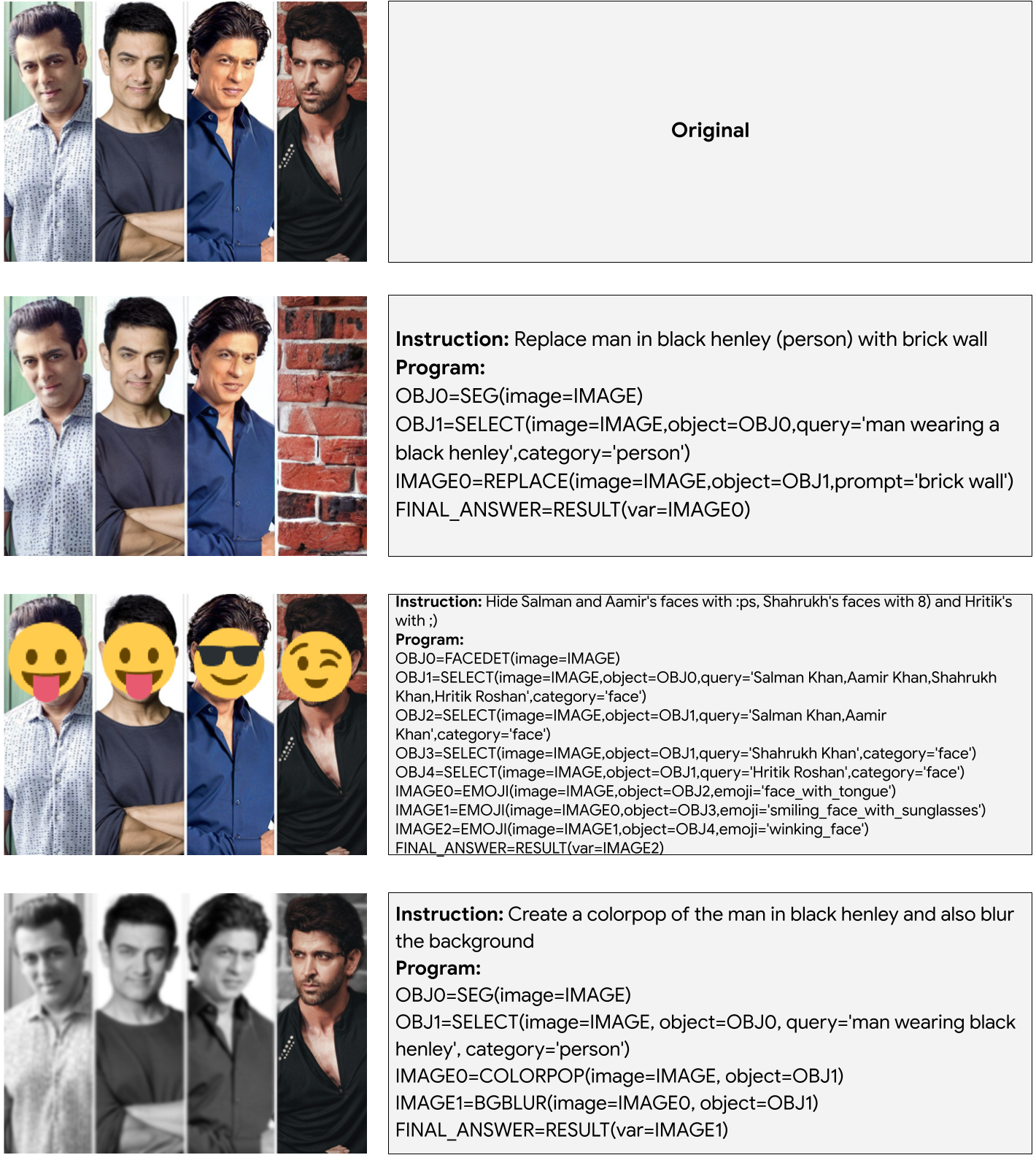}
  \caption{Image editing examples by zero-shot prompting \chatgpt with tool docs. Zero-shot prompting with docs is able to reproduce the results achieved by VisProg using few-shot demos~\cite{gupta2022visual}.}
  \label{fig:visprog_editing_examples}
\end{figure}

\paragraph{Image editing.}
We provide more image editing examples achieved by zero-shot prompting with tool docs in Figure~\ref{fig:visprog_editing_examples}. In particular, we show that with tool docs, we are able to reproduce the image editing examples achieved by VisProg~\cite{gupta2022visual} without using any few-shot demos, wherein VisProg relies on $10$ task-specific few-shot demos.

\end{document}